\documentclass[letterpaper, 10 pt, conference]{ieeeconf}  

\IEEEoverridecommandlockouts                              

\usepackage{graphics,graphicx} 
\graphicspath{{../Figures/}}
\usepackage{amsmath} 
\usepackage{amssymb}  
\usepackage{xcolor}

\usepackage{amsmath} 
\usepackage{amssymb}  
\usepackage{xcolor}
\usepackage{algorithm}
\usepackage{algpseudocode}
\usepackage{cite}
\usepackage{dblfloatfix}
\usepackage{mathtools} 
\usepackage{dutchcal} 
\usepackage{caption}
\usepackage{subcaption}
\usepackage{hyperref}
\usepackage{lipsum}
\usepackage{array} 

\usepackage{scalerel}
\newcommand{\norm}[1]{\left\lVert#1\right\lVert}

\newcommand{\x}{\mathbf{x}}

\newcommand{\f}{\mathbf{f}}

\newcommand{\fxx}{\mathbf{f}_{\mathbf{_{xx}}}}
\newcommand{\fuu}{\mathbf{f}_{\mathbf{{uu}}}}
\newcommand{\fux}{\mathbf{f}_{\mathbf{{ux}}}}

\newcommand{\q}{\mathbf{q}}
\newcommand{\U}{\mathbf{U}}

\newcommand{\s}{\mathbf{s}}

\newcommand{\Q}{\mathbf{Q}}
\newcommand{\R}{\mathbf{R}}
\newcommand{\dx}{\delta\x}
\newcommand{\du}{\delta\u}
\newcommand{\dX}{\delta\X}
\newcommand{\dU}{\delta\U}

\newcommand{\X}{\mathbf{X}}
\newcommand{\A}{\mathbf{A}}
\newcommand{\B}{\mathbf{B}}

\newcommand{\K}{\mathbf{K}}

\renewcommand{\S}{\mathbf{S}}

\renewcommand{\u}{\mathbf{u}}
\renewcommand{\P}{\mathbf{P}}
\renewcommand{\S}{\mathbf{S}}
\renewcommand{\u}{\mathbf{u}}

\renewcommand{\d}{\mathbf{d}}
\renewcommand{\d}{\mathbf{d}}
\renewcommand{\r}{\mathbf{r}}
\newcommand{\part}{\partial}
\renewcommand{\xi}{\x^{[i]}}

\newcommand{\optdu}{\delta\Tilde{\u}}

\newcommand{\trans}[1]{{#1}^{\top}}

\DeclareMathOperator*{\EC}{EC}

\usepackage{multirow}

\title{\LARGE \bf
A Unified Perspective on Multiple Shooting In Differential Dynamic Programming
}

\author{He Li$^{1}$, Wenhao Yu$^{2}$, Tingnan Zhang$^{2}$, and Patrick M. Wensing$^{1}$ 
\thanks{$^{1}$ He Li and Patrick M. Wensing are with the University of Notre Dame, IN, USA (\tt\small hli25@nd.edu, pwensing@nd.edu)}
\thanks{$^{2}$ Wenhao Yu and Tingnan Zhang are with Robotics at Google, Mountain View, CA, USA (\tt\small magicmelon@google.com, tingnan@google.com)}
}

\begin{document}

\maketitle
\thispagestyle{empty}
\pagestyle{empty}

\begin{abstract}
 Differential Dynamic Programming (DDP) is an efficient computational tool for solving nonlinear optimal control problems. It was originally designed as a single shooting method and thus is sensitive to the initial guess supplied. This work considers the extension of DDP to multiple shooting (MS), improving its robustness to initial guesses. A novel derivation is proposed that accounts for the defect between shooting segments during the DDP backward pass, while still maintaining quadratic convergence locally. The derivation enables unifying multiple previous MS algorithms, and opens the door to many smaller algorithmic improvements. A penalty method is introduced to strategically control the step size, further improving the convergence performance. An adaptive merit function and a more reliable acceptance condition are employed for globalization. The effects of these improvements are benchmarked for trajectory optimization with a quadrotor, an acrobot, and a manipulator. MS-DDP is also demonstrated for use in Model Predictive Control (MPC) for dynamic jumping with a quadruped robot, showing its benefits over a single shooting approach. Video \href{https://youtu.be/RNzE87sAn9E}{link}.
\end{abstract}



\section{Introduction} 
Model Predictive Control (MPC) is a powerful technique for controlling complex systems and has been widely used for many robotic systems, including quadrotors \cite{falanga2018pampc}, quadruped robots \cite{di2018dynamic, grandia2019feedback}, and humanoid robots \cite{kuindersma2016optimization}. MPC needs to efficiently and reliably solve a sequence of finite horizon optimal control problems (OCPs) of the form
\begin{subequations}\label{eq:OCP}
\begin{IEEEeqnarray}{rc}
\min_{\u(\cdot)} \ \ \ \ & \int_0^T \ell_c(\x(t), \u(t) ) {\rm d}t + \phi(\x(T)) \label{eq:ct_cost}\\
\text{subject~to} \ \ \ \ & \dot{\x} = \f_c(\x, \u) \label{eq:ct_dyn}
\end{IEEEeqnarray}    
\end{subequations}
where $T$ is the prediction horizon, $\x$ the state variable, $\u$ the control variable, $\ell_c$ the running cost, $\phi$ the terminal cost, and $\f_c$ the dynamics function. The problem~\eqref{eq:OCP} is an infinite-dimensional optimization problem, as it is in continuous time, and the dynamics are highly nonlinear for many robotics systems. Therefore, an analytical solution, in general, does not exist, and numerical methods are often employed.
 One commonly used class of approaches are \textit{direct methods}. A \textit{direct method} parameterizes the state and control using a finite number of variables, and transcribes the original OCP~\eqref{eq:OCP} into a nonlinear optimization problem. The problem transcription most often takes one of three approaches, \textit{single shooting} (SS), \textit{multiple shooting} (MS), and \textit{collocation}. This work focuses on the first two approaches.

Differential Dynamic Programming (DDP) \cite{mayne1966second} is a \textit{single shooting} method to solve the OCP~\eqref{eq:OCP}, and is shown to have a local quadratic convergence rate \cite{liao1992advantages}. It naturally exploits the temporal structure of the transcribed problem \eqref{eq:OCP} by successively solving a sequence of sub-problems, resulting in linear computational cost with horizon length $T$. These sub-problems originate from locally solving Bellman's equation, which additionally gives a local feedback policy without additional computation cost \cite{grandia2019feedback}. These properties make DDP exceptionally well suited for MPC with a long prediction horizon. In the past decade, DDP and its variants have been widely used for MPC of quadruped locomotion \cite{grandia2019feedback,li2022versatile, mastalli2022inverse}, humanoid balancing \cite{koenemann2015whole, dantec2022whole}, and loco-manipulation \cite{bjelonic2021whole}. The downside of DDP, common with many other single-shooting algorithms, is that an initial guess for the state trajectory cannot be directly supplied, and it is sensitive to the initial guess of the control trajectory \cite{bock1984multiple, diehl2006fast}.

\begin{figure}[t]
    \centering
    \includegraphics[width = 0.9\linewidth]{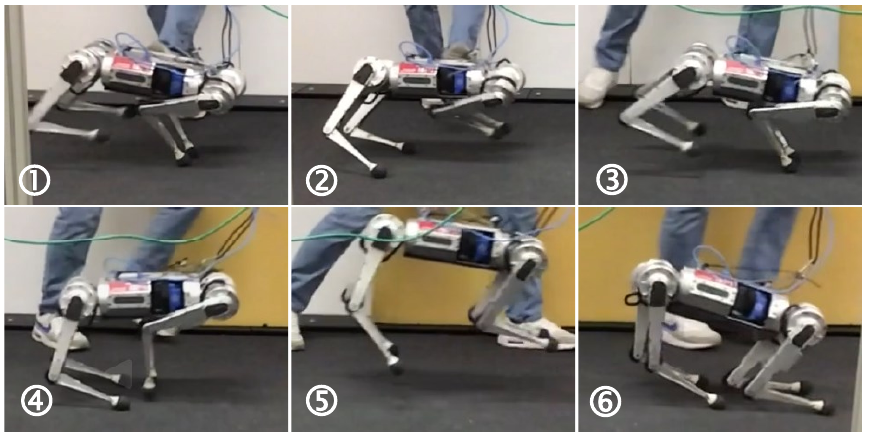}
    \caption{Time-series snapshots of robust quadruped jumping enabled by the proposed MS-DDP algorithm within an MPC controller. 1. The robot is pushed before jumping. 2. Recovering stability. 3. Back to bounding (1 m/s). 4. Jumping starts. 5. Middle of the jump. 6. Jumping ends.}
    \label{fig:hardware-snap}
\end{figure}
\begin{figure}[b]
    \centering
    \includegraphics[width = 0.95\linewidth]{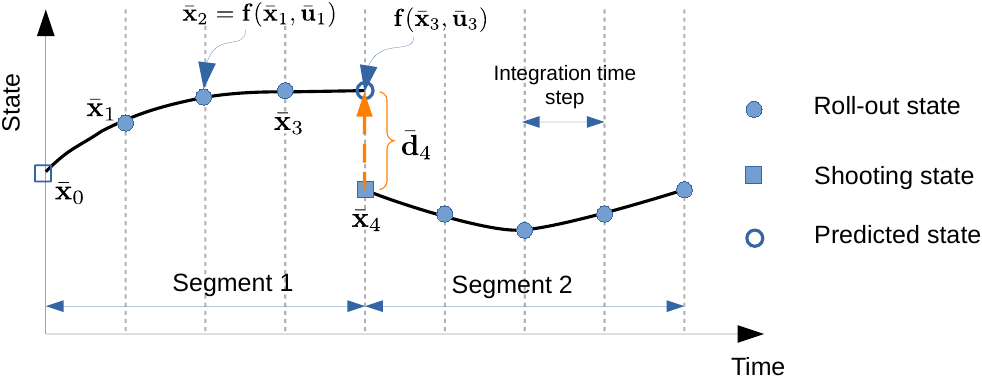}
    \caption{Illustration of multiple shooting for a nominal trajectory $\big(\Bar{\x},\Bar{\u}\big)$. The shooting state is a decision variable that can be initialized and updated. The roll-out state is always overwritten by dynamics simulation propagated from a previous time instant.}
    \label{fig:ms_illustration}
\end{figure}

Multiple shooting (Fig.~\ref{fig:ms_illustration}) \cite{bock1984multiple, diehl2006fast} alleviates the sensitivity problem of single shooting by introducing intermediate state variables (also known as \textit{shooting states}). The shooting states divide a trajectory into several sub-intervals, known as shooting segments. Numerical integration is performed separately on each segment using the shooting state as an initial condition. Continuity constraints that link the shooting state to a previous segment need to be satisfied at convergence, so that the resulting trajectory is dynamically feasible. The problem sensitivity is distributed across the segments at the price of potential discontinuity (also known as a defect) at each shooting state. This work proposes a new derivation for addressing defects in DDP, which unifies past approaches.

Previous works \cite{giftthaler2018family, Howell19, mastalli2020crocoddyl, pellegrini2020multiple} on pairing MS and DDP mainly differ in the way the continuity constraints are handled, and can be categorized into two classes. The first class \cite{Howell19, pellegrini2020multiple} employs Augmented Lagrangian (AL) methods. This approach exhibits low accuracy, but can be addressed using a projection step \cite{Howell19} or a second-order method for the update of the Lagrange multiplier \cite{pellegrini2020multiple}. The second class is inspired by Sequential Quadratic Programming (SQP) \cite{giftthaler2018family, mastalli2020crocoddyl}, which successively linearizes the constraints and solves a quadratic problem. This approach shows good convergence near the (local) solution, but care must be taken when the initial guess is far away from the solution. Globalization strategies such as employing filter-based line search methods \cite{grandia2022perceptive} or merit functions \cite{jallet2022constrained, singh2022optimizing} can address this problem. 

\subsection{Contributions}
The contributions of this paper are as follows. First, we provide a novel derivation that naturally embeds the multiple-shooting formulation into the DDP algorithm. This derivation enables incorporating second-order sensitivity information across shooting segments, resulting in local quadratic convergence. Further, the derivation enables unifying multiple previous algorithms \cite{giftthaler2018family, mastalli2020crocoddyl}, and opens the door to broadly applicable algorithmic improvements. Second, several algorithmic enhancements are introduced to improve the algorithm's robustness, including a penalty method, a more reliable acceptance condition, and an adaptive merit function. We numerically benchmark the proposed enhancements together with previous algorithms \cite{giftthaler2018family, mastalli2020crocoddyl} in terms of robustness and convergence rate on several robotic systems. Finally, we implement the algorithm for Model Predictive Control (MPC) of a quadruped robot, and demonstrate its superior performance to single-shooting DDP (SS-DDP).
\section{Multiple-Shooting Formulation of Optimal Control Problems} 
Multiple shooting transcribes the continuous-time OCP~\eqref{eq:OCP} to a discrete-time OCP 
\begin{subequations}\label{eq:dt_opt}
\begin{IEEEeqnarray}{rl}
\min_{\U, \X} \ \ & J(\X, \U) = \sum_{k=0}^{N-1} \ell_k(\x_k, \u_k) + \phi(\x_N)\label{eq:dt_cost} \\
\text{subject~to} \ \ & \underbrace{\f(\x_k, \u_k)-\x_{k+1}}_{\d_{k+1}(\x_k,\u_k,\x_{k+1})}  =  0 \label{eq:dt_dyn}
\end{IEEEeqnarray}
\end{subequations}
where $0 \leq k \leq N$ denotes time instant with $t_N = T$, $\U = \{\u_k\}_{k=0}^{N-1}$, $\X = \{\x_k\}_{k=0}^N$, $\ell_k(\x_k, \u_k)$ the discretized running cost function, $\f(\x_k, \u_k)$ the discretized dynamics function, and $\d_{k+1}(\x_k,\u_k,\x_{k+1})$ the defect function that measures the dynamics feasibility of a triplet $(\x_k,\u_k,\x_{k+1})$. The formulation~\eqref{eq:dt_opt} considers both the control and state as decision variables. We follow the hybrid scheme in \cite{giftthaler2018family} to categorize the state variables to two types, namely a shooting state and a roll-out state. If a state variable $\x_{k+1}$ is a shooting state, then it can be independently initialized and updated, thus the defect function $\d_{k+1}$ is not necessarily zero until the algorithm converges. If a state variable is a roll-out state, then it is always overwritten by the dynamics simulation propagated from a previous time step, thus the defect function is identically equal to zero. 

Figure~\ref{fig:ms_illustration} further illustrates these definitions. A trajectory of length $N-1$ is evenly divided into $M$ segments, each incorporating $m = (N-1)/M$ time steps. The initial state of each segment (i.e., at time index $k={m, 2m,\cdots,(M-1)m}$) comprises the set of shooting states, with time indices denoted $\mathbb{I}$, whereas the complementary set $\Tilde{\mathbb{I}} = \{0,\cdots,N\} \setminus \mathbb{I}$ denotes the indices of roll-out states. Given a nominal trajectory $\big(\Bar{\X},\Bar{\U}\big)$, dynamics simulation is performed on each segment using the corresponding shooting state as the initial condition. If  $k+1 \in \Tilde{\mathbb{I}}$, then $\Bar{\x}_{k+1}$ is overwritten by $\f(\Bar{\x}_k, \Bar{\u}_k)$. Otherwise, $\Bar{\x}_{k+1}$ remains unchanged, and the defect $\Bar{\d}_{k+1} = \d_{k+1}(\Bar{\x}_{k}, \Bar{\u}_k,\Bar{\x}_{k+1})$ is measured. We denote the nominal value of a variable with $\Bar{\cdot}$ overhead throughout the paper. The optimization algorithms developed in this work for problem~\eqref{eq:dt_opt} will need to drive the defects to zero while minimizing the cost function.
\section{Differential Dynamic Programming For Multiple-Shooting OCP}\label{sec:msddp}
DDP applies a local version of Bellman's principle of optimality to the OCP~\eqref{eq:dt_opt} by considering a small perturbation $\big(\dX, \dU\big)$ near the nominal trajectory $\big(\Bar{\X},\Bar{\U}\big)$
\begin{equation}\label{eq:delta_bellman}
    v_k(\dx_k) = \min_{\du_k} \big(\underbrace{\delta\ell_k(\dx_k, \du_k) + v_{k+1}(\dx_{k+1})}_{Q_k(\dx_k,\du_k)} \big)
\end{equation}
where $v_k(\cdot)$ denotes the local value function, $\delta\ell_k(\dx_k,\du_k)$ quadratically approximates the perturbation of $\ell_k(\x_k,\u_k)$ at ($\Bar{\x}_k,\Bar{\u}_k$). We use $Q_k(\cdot,\cdot)$ to denote the Bellman objective of eq.~\eqref{eq:delta_bellman} for simplicity. In the traditional case of SS-DDP, the algorithm considers every state as a roll-out state, thus the perturbation $\big(\dX, \dU\big)$ needs to satisfy
\begin{equation}\label{eq:pertub_dyn_1}
    \dx_{k+1} = \f(\Bar{\x}_k+\dx_k, \Bar{\u}_k+\du_k) - \f(\Bar{\x}_k,\Bar{\u}_k).
\end{equation}
Substituting eq.~\eqref{eq:pertub_dyn_1} to the local Bellman's equation~\eqref{eq:delta_bellman} and approximating $\f$ quadratically results in standard Ricatti-like difference equations and a local policy \cite{mayne1966second, tassa2012synthesis}. 
\subsection{Backward Sweep Accounting For Defect}
Traditional DDP, however, is not applicable to deal with the shooting state as introduced in the formulation~\eqref{eq:dt_opt} because of the defect. In this work, we revise the traditional DDP backward sweep to account for the defect using a simple but effective trick. Eq.~\eqref{eq:pertub_dyn_1} is modified as below
\begin{equation}\label{eq:pertub_dyn_2}
    \dx_{k+1} = \f(\Bar{\x}_k+\dx_k, \Bar{\u}_k+\du_k) - \Bar{\x}_{k+1}.
\end{equation}
The equation~\eqref{eq:pertub_dyn_2} allows for dynamic infeasibility in the initial guess or intermediate trajectories ($\Bar{\X}, \Bar{\U}$). 
To derive the update policy associated with~\eqref{eq:pertub_dyn_2}, we approximate $v_k(\dx_k)$ to the second order as in DDP
\begin{equation}\label{eq:dV}
    v_k(\dx_k) = \frac{1}{2}(\dx_k)^\top\S_k\dx_k + \s_k^\top\dx_k + s_k.
\end{equation}
where $\S_k$, $\s_k$, $s_k$ are the Hessian, gradient, and zero-order terms of $v_k(\dx_k)$. By approximating $\f$ in eq.~\eqref{eq:pertub_dyn_2} to the second order and substituting~\eqref{eq:dV} into eq.~\eqref{eq:delta_bellman}, we get 
\newcommand{\pmwspace}{.5ex}
\begin{equation}\label{eq:Qapprox}
    Q_k(\delta\x, \delta\u) \approx
    \frac{1}{2}
    \begin{bmatrix}
    1 \\[\pmwspace] \delta\x \\[\pmwspace] \delta\u
    \end{bmatrix}^T
    \begin{bmatrix}
        0   &   \Q_{\x,k}^T   &   \Q_{\u,k}^T \\[\pmwspace]
        \Q_{\x,k}&   \Q_{\x\x,k}    &   \Q_{\u\x,k}^T \\[\pmwspace]
        \Q_{\u,k} &   \Q_{\u\x,k}    &   \Q_{\u\u,k}   \end{bmatrix}
     \begin{bmatrix}
    1 \\[\pmwspace] \delta\x \\[\pmwspace] \delta\u
    \end{bmatrix},
\end{equation}
where
\begin{subequations}\label{eq:Qs_DDP}
\begin{align}
    \Q_{\x,k} &= \q_k + \A_k^{\top}(\s_{k+1} + \color{blue}\S_{k+1}\Bar{\d}_{k+1}), \label{eq:Qs_DDP_Qx}\\
    \Q_{\u,k} &= \r_k + \B^{\top}_k(\s_{k+1} + \color{blue}\S_{k+1}\Bar{\d}_{k+1}), \label{eq:Qs_DDP_Qu}\\
    \Q_{\x\x,k} &= \Q_k + \A^{\top}_k\S_{k+1}\A_k + \color{red}\s_{k+1}\cdot\fxx{}_{,k}, \label{eq:Qs_DDP_Qxx}\\
    \Q_{\u\u,k} &= \R_k + \B^{\top}_k\S_{k+1}\B_k + \color{red}\s_{k+1}\cdot\fuu{}_{,k}, \label{eq:Qs_DDP_Quu} \\
    \Q_{\u\x,k} &= \P_k + \B^{\top}_k\S_{k+1}\A_k + \color{red}\s_{k+1}\cdot\fux{}_{,k}, \label{eq:Qs_DDP_Qux}
\end{align}
\end{subequations}
in which $\A_k = \left.\frac{\part \f}{\part \x}\right|_{(\Bar{\x}_k,\Bar{\u}_k)}$, $\B_k = \left. \frac{\part \f}{\part \u}\right|_{(\Bar{\x}_k, \Bar{\u}_k)}$, $\q_k$ and $\r_k$ are gradients of $\ell_k$ w.r.t. $\x$ and $\u$ respectively, $\Q_k$, $\R_k$, and $\P_k$ are second-order partials of $\ell_k$, and $\f_{(\cdot,\cdot)}$ are second-order partials of $\f$. The recursive equations for $\S_k$, $\s_{k}$, and $s_k$ are
\begin{subequations}\label{eq:updateV}
\begin{align}
\S_k &= \Q_{\x\x,k} - \Q_{\u\x,k}^T \Q_{\u\u,k}^{-1} \Q_{\u\x,k} \label{eq_Vxx}\\
    \s_k &= \Q_{\x,k} - \Q_{\u\x,k}^T\Q_{\u\u,k}^{-1}\Q_{\u,k}, \label{eq_Vx}\\
    s_k &= s_{k+1}  - \frac{1}{2}\Q_{\u,k}^T \Q_{\u\u,k}^{-1} \Q_{\u,k} + \nonumber \\ 
    &\quad\quad\quad\quad\quad\quad\quad\quad \color{blue} \trans{\s}_{k+1}\Bar{\d}_k + \frac{1}{2}\trans{\Bar{\d}}_{k+1}\S_{k+1}\Bar{\d}_{k+1},
    \label{eq_deltaV}   
\end{align}
\end{subequations}
with the boundary conditions $\S_N = \Q_N, \s_N = \q_N, s_N = 0$. Minimizing~\eqref{eq:Qapprox} over $\du_k$ results in a local optimal control policy 
\begin{equation}
    \du^*_k = \optdu_k + \K_k\dx_k ,
\end{equation}
where
\begin{equation}\label{eq:local_opt_ctrl}
    \optdu_k = -\Q_{\u\u,k}^{-1}\Q_{\u,k}, \quad
    \K_k =  -\Q_{\u\u,k}^{-1}\Q_{\u\x,k}.
\end{equation}
 The equations \eqref{eq:Qs_DDP} and \eqref{eq:updateV} provide general formulas for the backward sweeps of four DDP variants
 \begin{enumerate}
    \item \textit{MS-DDP} (this work): Multiple shooting, second-order.
     \item \textit{SS-DDP}\cite{mayne1966second, nganga2021accelerating}: Single shooting, second-order (no blue).
     \item \textit{MS-iLQR}\cite{giftthaler2018family,mastalli2020crocoddyl}: Multiple shooting, first-order (no red).
     \item \textit{SS-iLQR} \cite{tassa2012synthesis, farshidian2017efficient}: Single shooting, first-order (no blue and no red).   
\end{enumerate}
We will investigate the effect of the second-order dynamics on local convergence in Section~\ref{subsec:converge}.

\subsection{Forward Roll-out}
The control policy~\eqref{eq:local_opt_ctrl} provides a search direction for the control update from iteration to iteration. A forward roll-out for the dynamics must be conducted to obtain a search direction for the state update. In this section, we explore three methods for the forward roll-out that mainly differ in the dynamics, namely a \textit{linear roll-out} \cite{giftthaler2018family, jallet2022implicit}, \textit{nonlinear roll-out} \cite{mastalli2020crocoddyl}, and \textit{hybrid roll-out} \cite{giftthaler2018family}. For clarity, we repeat some notations here, $(\Bar{\X}, \Bar{\U})$ denotes the nominal trajectory, and $(\X', \U')$ denotes the new trajectory. Whichever roll-out method is used, the control update always has the same format
\begin{equation}\label{eq:control_update}
    \u'_k = \Bar{\u}_k + \underbrace{\alpha\optdu_k + \K_k(\x'_k - \Bar{\x}_k)}_{\du_k(\alpha)}.
\end{equation}
where $\x'_0 = \Bar{\x}_0$, $\alpha \in (0,1]$ is the step size, which is used in backtracking line search for global convergence \cite{tassa2012synthesis}, and $\du_k(\alpha)$ is the scaled search direction. The main difference between the three roll-out methods is in the state update.

A \textit{linear roll-out} simulates the linearized dynamics of \eqref{eq:pertub_dyn_2} using the control policy~\eqref{eq:local_opt_ctrl} 
\begin{equation}\label{eq:linear_rollout}
    \x'_{k+1} = \Bar{\x}_{k+1} + \underbrace{[\A_k(\x'_k - \Bar{\x}_k) + \B_k\du_k(\alpha)] + \alpha\Bar{\d}_{k+1}}_{\delta\x_{k+1}(\alpha)}
\end{equation}
where $\dx_{k+1}(\alpha)$ is the scaled search direction for $\Bar{\x}_{k+1}$. Note that $\dx_{k+1}(\alpha)$ scales linearly with $\alpha$ in this case. This method is computationally cheap, since it only needs to be executed once with $\alpha=1$, and a line search can then be performed in parallel across all time instants. The downside, however, is that the method simplifies the nonlinear dynamics, and thus is subject to prediction error \cite{mastalli2020crocoddyl}. Further, the linear roll-out requires that every state is a \textit{shooting state}, losing the flexibility for other algorithmic options. 

To account for nonlinearity, a \textit{nonlinear roll-out} \cite{mastalli2020crocoddyl} simulates the original nonlinear dynamics using the control policy~\eqref{eq:local_opt_ctrl}
\begin{equation}\label{eq:nonlinear_ls}
    \x'_{k+1} = \Bar{\x}_{k+1} + \underbrace{[\f(\x'_{k}, \u'_k)  - \f(\Bar{\x}_{k}, \Bar{\u}_k)] + \alpha \Bar{\d}_{k+1}}_{\delta\x_{k+1}(\alpha)}
\end{equation}
The nonlinear roll-out~\eqref{eq:nonlinear_ls} avoids the linear prediction error, with the above equivalent to the scheme in \cite{mastalli2020crocoddyl}. The derivation here, however, more resembles the behavior of \eqref{eq:linear_rollout} with a replacement of nonlinear dynamics. Unlike the linear roll-out, which enables parallel computation, the nonlinear roll-out has to be performed serially, thus potentially hindering the computational performance.

A \textit{hybrid roll-out} method was proposed in \cite{giftthaler2018family}, which attempts to combine the benefit of both. Namely, the hybrid method first performs a linear roll-out~\eqref{eq:linear_rollout} to obtain $\delta\x_{k+1}(\alpha)$ for the shooting states. The updated shooting state is then used as an initial condition, and the nonlinear roll-out can be performed independently on each shooting segment for the roll-out states. With this method, the search direction for the shooting nodes only needs to be computed once, and the line search can then be performed in parallel on each shooting segment. Previous work \cite{giftthaler2018family} has shown that the hybrid roll-out exhibits better global convergence than the linear roll-out, thus the linear roll-out is not considered here.

The unified MS-DDP framework developed in this work synthesizes all four DDP variants for the backward pass, and the nonlinear and hybrid forward roll-out. The multiple algorithm configurations enabled by this synthesis may produce different convergence behaviors as will be shown in Section~\ref{subsec:converge}.

\section{Improving Robustness and Flexibility}\label{sec:globalization}
The previous section introduced methods for computing the search direction that updates the control and state variables. This section presents new techniques that better measure the quality of a search step to determine the step size. Further, we introduce an advance that alters the search direction by modifying the backward sweep (eq.~\eqref{eq:updateV}).

\subsection{Merit Function}
MS-DDP needs to balance two goals, minimizing the cost function~\eqref{eq:dt_cost} while decreasing the defects~\eqref{eq:dt_dyn}. These two goals can sometimes be conflicting with each other since reducing the defect may otherwise increase the cost \cite{nocedal2006numerical}. A merit function synthesizes the two objectives into one single function, and is widely used in Nonlinear Programming (NLP) for constrained optimization \cite{nocedal2006numerical, gill2005snopt}. In this work, we use a merit function to monitor the progress of MS-DDP. The merit function considered is an $L_p$-norm merit function since it has been proven to be exact \cite{nocedal2006numerical} and does not require an estimate of the Lagrange multiplier. The $L_p$-norm merit function for problem~\eqref{eq:dt_opt} is defined as 
\begin{equation}\label{eq:merit}
    M(\X,\U) = J(\X, \U) + \mu \norm{\d(\X,\U)}_p
\end{equation}
where $\d(\cdot, \cdot)$ is a vector aggregating all the defects, $\mu > 0$ is a weighting parameter that balances the cost function and the defects violation. To avoid meticulous tuning of $\mu$ for different problems, we consider an adaptive scheme that is motivated by \cite{nocedal2006numerical} so that $\mu$ is updated as
\begin{equation}\label{eq:adaptive}
    \mu = \frac{\EC(\alpha)}{(1-\rho)\norm{\d(\Bar{\X},\Bar{\U})}_p} + \mu_{0} ~~ {\textrm{when}} ~~ \norm{\d(\Bar{\X},\Bar{\U})}_p > \kappa_d
\end{equation}
where $\mu_0>0$ sets a safety margin, $\EC(\alpha)$ denotes the expected cost change due to $(\dX(\alpha),\dU(\alpha))$, which is discussed in the next subsection, $0<\rho<1$ is a fixed tuning parameter, and $\kappa_d>0$ is the threshold for updating $\mu$. We use $\rho = 0.5$, and $\mu_0 = 10$ for all problems in this work without further tuning. An alternative approach to using a merit function is a filter-based technique \cite{pavlov2021interior, grandia2022perceptive}. 

\subsection{Acceptance Condition}
A simple condition to accept a search step $(\dX(\alpha),\dU(\alpha))$ is to ensure the merit function~\eqref{eq:merit} is decreased, i.e., $M(\X',\U') - M(\Bar{\X}, \Bar{\U}) < 0$. This requirement, however, may not produce convergence to a local optimum, as shown in \cite{nocedal2006numerical}. In this work, we use an Armijo condition to impose sufficient merit reduction 
\begin{equation}\label{eq:armijo}
    M(\X',\U') < M(\Bar{\X}, \Bar{\U}) + \gamma\big(\EC(\alpha) - \alpha\mu\norm{\d(\Bar{\X},\Bar{\U})}_p \big)
\end{equation}
where $0 < \gamma < 1$ is a tuning parameter, $\EC(\alpha)$ is the expected cost change, and 
\begin{equation}\label{eq:EC}
    \EC(\alpha) = \alpha\text{EC}_1 + \frac{1}{2}\alpha^2\text{EC}_2
\end{equation}
in which
\begin{subequations}
\begin{align}
 \label{eq:EC1}
    \scaleobj{.92}{\rm{EC}_1} &\scaleobj{.92}{=}\, \scaleobj{.92}{\q^{\top}_N\dx^l_N + \sum_{k=0}^{N-1}\q^{\top}_k\dx^l_k + \r^{\top}_k\du^l_k,} \\
  \scaleobj{.92}{\rm{EC}_2} &\scaleobj{.92}{=}\, \scaleobj{.92}{\dx^{l\top}_N\Q_N\dx^l_N + \sum_{k=0}^{N-1} \dx^{l\top}_k\Q_k\dx^l_k + \du^{l\top}_k\R_k\du^l_k} \nonumber\\
  \scaleobj{.92}{}&{\qquad\quad\qquad\qquad\qquad\qquad + \du^{l\top}_k\P_k\dx^l_k} 
\end{align}
\end{subequations}
where $\du^l_k$ and $\dx^l_k$ are obtained via the linear roll-out~\eqref{eq:control_update} and \eqref{eq:linear_rollout}. The EC~\eqref{eq:EC} provides an exact cost change in the case of linear dynamics and quadratic cost approximation. It is equivalent to the model of expected cost change \cite{tassa2012synthesis} computed from the iLQR backward pass if all defects are zero. We find that this expectation model further improves upon the one used within \cite{mastalli2020crocoddyl, FDDP-Derivation}, in the sense that cost effects from the defect are fully treated, leading to an exact match when applied to linear quadratic MS problems. 

\subsection{Modified Backward Sweep Using A Penalty Method}
A backtracking line search is employed to determine the step size $\alpha$. The step size $\alpha$, however, may still be so small that the global convergence is hindered. In this section, we introduce a penalty method to improve this process. Note that in Fig.~\ref{fig:ms_illustration} the shooting state adjusts the defect size only from the right side, but the rolled-out state $\x^-_{k+1}:= \f(\x_k,\u_k)$ on the left side is not aware of this change. The proposed method strategically controls the defect from both sides, by imposing a penalty that promotes connecting the segments from the left side as well. This is done by adding the penalty term $\norm{\x^-_{k+1} - \Bar{\x}_{k+1}}^2_{\Q_{d_{k+1}}}$ to the cost function in~\eqref{eq:dt_opt}, where $\Q_{d_{k+1}}$ is a positive definite weight matrix if $k+1$ is a shooting state, and is zero if $k+1$ is a roll-out state. Adding the penalty term to~\eqref{eq:dt_opt} amounts to modifying the backward sweep equations~\eqref{eq:Qs_DDP_Qu} and ~\eqref{eq:Qs_DDP_Qx} such that 
\begin{subequations}
    \begin{align}
        \s_{k+1} &\leftarrow \s_{k+1} - \Q_d\Bar{\d}_k \\
        \S_{k+1} &\leftarrow \S_{k+1} + \Q_d
    \end{align}
\end{subequations}
This penalty method is similar in spirit to the proximal term \cite{bambade2022prox} that helps improve the conditioning of the KKT system.
The benefit of adding the terminal cost to the algorithm performance will be demonstrated in the result section.

\section{Discussion on The Unified Perspective}
The search direction computation in Section~\ref{sec:msddp} and the globalization method for step acceptance in Section~\ref{sec:globalization} comprise one iteration of the MS-DDP algorithm. A brief summary of the overall MS-DDP framework is given below
\begin{enumerate}
    \item Give the nominal trajectory $(\Bar{\x},\Bar{\u})$, optimization horizon $N$, and number of shooting segments $M$.
    \item Perform dynamics approximation and cost function approximation around the nominal trajectory.
    \item Compute the optimal control policy~\eqref{eq:local_opt_ctrl} using one of the four DDP variants.
    \item Perform backtracking line search using either hybrid roll-out or nonlinear roll-out, the adaptive merit function~\eqref{eq:adaptive} and the acceptance condition~\eqref{eq:armijo}.
    \item Check the cost convergence criterion and feasibility. Go to 2) if not satisfied, and terminate otherwise.
\end{enumerate}

The MS-DDP framework herein provides a unified perspective since it enables multiple configurations, as summarized in Table~\ref{tab:unified_msddp}. Four DDP variants could be used in performing the backward sweep, and two forward roll-out methods could be employed for line search. Further, the penalty method and the adaptive merit function offer additional options for improving the algorithm robustness and flexibility. As we will show in the next section, the performance of an algorithm to solve \eqref{eq:dt_opt} is problem-dependent. The unified MS-DDP framework enables easy comparison across different algorithm configurations. In fact, if we choose the MS-iLQR for computing the backward sweep (eq.~\eqref{eq:Qs_DDP} and eq.~\eqref{eq:updateV}), the hybrid roll-out for line search, and $\gamma=0$ for the acceptance condition, then the proposed framework can be simplified to GN-iLQR \cite{giftthaler2018family}. If we choose the MS-iLQR for the backward sweep, the nonlinear roll-out for line search, and compute $\EC$ with an approximate model, the MS-DDP framework is then simplified to FiLQR \cite{mastalli2020crocoddyl}\footnote{For consistency with the nomenclature in this paper, we depart from the name ``FDDP'' used in \cite{mastalli2020crocoddyl}  and use FiLQR instead, which reflects the exclusive use of first-order dynamics sensitivities in the method.}. 
\begin{table}[t]
    \centering
    \caption{Summary of algorithmic configurations with the unified MS-DDP}
    \begin{tabular}{c|c|c|c}
    \hline
        & Unified MS-DDP & FiLQR \cite{mastalli2020crocoddyl} & GN-iLQR \cite{giftthaler2018family} \\
    \hline
     Backward  & SS, MS & SS, MS & SS, MS \\
    Sweep Dyn~\eqref{eq:Qs_DDP} 
     & $1^{\text{st}}$ order & $1^{\text{st}}$ order & $1^{\text{st}}$ order\\
     & $2^{\text{nd}}$ order &                    &\\     
     \hline
     Forward  & Hybrid  & Nonlinear & Hybrid \\
    Roll-out & Nonlinear & & \\
     \hline
     Merit  & Adaptive & Cost function & Constant\\
    Function & Constant & & \\
                    & Cost function & & \\
     \hline
     Expectaion  & Exact & Approximate & Simple\\   
     Model       & Approximate & & \\
     \hline
    \end{tabular}    
    \label{tab:unified_msddp}
\end{table}
\section{Numerical Results}
The MS-DDP framework is benchmarked on three problems for numerical analysis. Each problem is associated with moving a robotic system from an initial state to a terminal state. A semi-implicit Euler method is used for integration, with the integration time step 0.02 s. Quadratic cost functions are used for all problems. We use the $L_2$-norm to measure the total defect. Algorithm convergence is approximately achieved if the normalized cost change is within $1e{-8}$, and the total defect is less than $1e{-3}$. We briefly describe each problem here.
\begin{enumerate}
    \item \textit{Acrobot}: A two-link manipulator where only the second joint is actuated. The acrobot needs to swing from a downward configuration up to an upward configuration in four seconds.
    \item \textit{Quadrotor}: The quadrotor is modeled as a rigid body with four thrust inputs, each at a certain distance from the center of mass \cite{li2021model}. The quadrotor is supposed to travel 5 m from one static position to another static position in four seconds.
    \item \textit{Manipulator}: The Kuka iiwa 7-DoF serial manipulator is used. The robot needs to swing from an upward configuration to a bending configuration in four seconds.
\end{enumerate}

\subsection{Numerical Convergence Analysis}\label{subsec:converge}
The local convergence rate of an algorithm is defined in the neighborhood of a local optimum, whereas global convergence is characterized by its capability to move a remote initial guess to a local optimum. In this section, we statistically quantify these properties for the MS-DDP framework configured with and without second-order dynamics.
\subsubsection{Local Convergence} 
\begin{figure}[t]
    \centering
    \includegraphics[width = \linewidth]{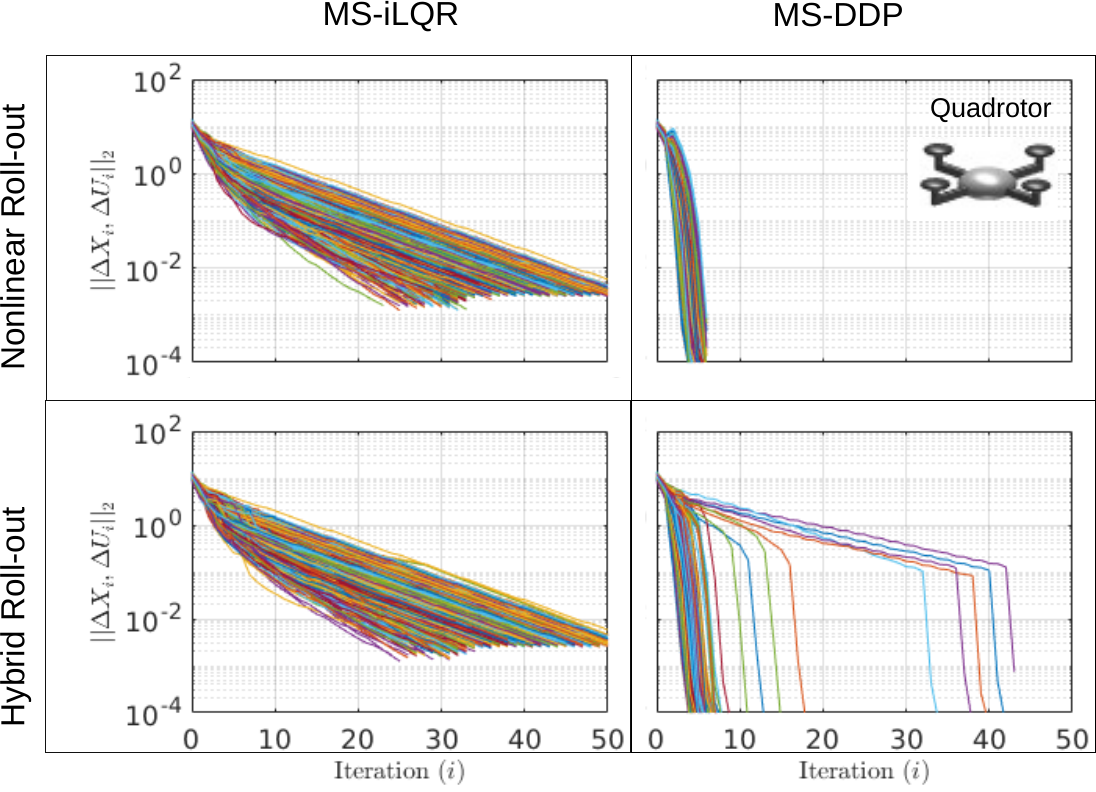}
    \caption{Local convergence for 1000 randomly sampled initial trajectories of the quadrotor in the neighborhood of the optimal trajectory. Left: MS-iLQR (Gauss-Newton Hessian approximation). Right: MS-DDP with full Hessian.}
    \label{fig:local_converg}
\end{figure}
Denote $(\X^*, \U^*)$ a local optimum of problem~\eqref{eq:dt_opt}, $(\X_j, \U_j)$ the $j^{th}$ iterate produced by the MS-DDP framework, and $(\Delta\X_j, \Delta\U_j)$ the difference between the $j^{th}$ iterate and the local optima $(\X^*, \U^*)$. The rate of local convergence is characterized by
\begin{equation}
    \lim_{j\rightarrow\infty} \norm{(\Delta\X_{j+1}, \Delta\U_{j+1})}_2 = \kappa \norm{(\Delta\X_j, \Delta\U_j)}_2^{\epsilon}
\end{equation}
where $\epsilon=1$ and $0<\kappa<1$ indicates linear convergence, and $\epsilon =2$ and $\kappa > 0$ indicates quadratic convergence. Four MS-DDP configurations are studied here: with and without the second-order dynamics in eq.~\eqref{eq:Qs_DDP}, and hybrid vs. nonlinear roll-out. All algorithms are benchmarked on the quadrotor problem. Each algorithm is configured to have 200 shooting segments, and is tested in a Monte Carlo fashion with 1000 initial guesses. The state guess is randomly sampled from a uniform distribution around $\X^*$ while the control guess remains the same as $\U^*$. Figure~\ref{fig:local_converg} depicts the local convergence results for all four algorithms. Linear convergence is always obtained if only first-order dynamics is used in (eq.~\eqref{eq:Qs_DDP}). Adding the second-order dynamics can improve the convergence rate. Quadratic convergence is consistently achieved over all samples with the nonlinear roll-out. For MS-DDP with the hybrid roll-out, the local convergence rate for certain samples is somewhere between linear convergence and quadratic convergence. This difference is reasonable in the sense that the shooting state update is based on the linearized dynamics with the hybrid roll-out, thus subject to prediction error, whereas it is based on nonlinear dynamics with the nonlinear roll-out, thus free of prediction error.

\subsection{Effects of Acceptance Condition}
\begin{figure}[b]
    \centering
    \includegraphics[width=\linewidth]{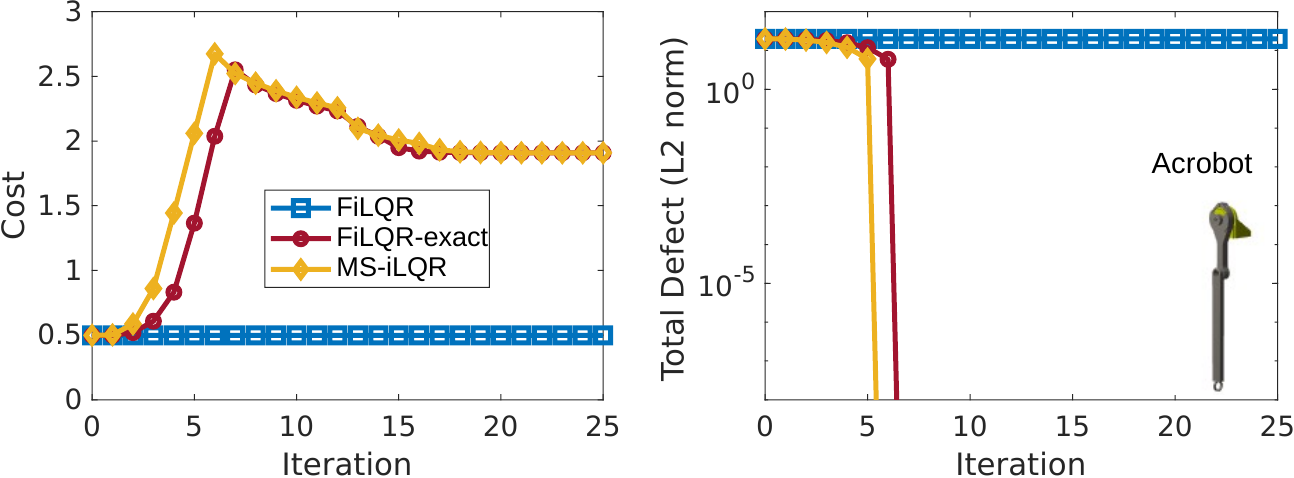}
    \caption{Effects of expected cost change on the global convergence for the acrobot with different algorithm configurations: FiLQR, FiLQR with exact expected cost change (eq.~\eqref{eq:EC}), MS-iLQR with adaptive merit function and nonlinear roll-out.}
    \label{fig:EC_compare}
\end{figure}
\begin{figure}[b]
    \centering
    \includegraphics[width = \linewidth]{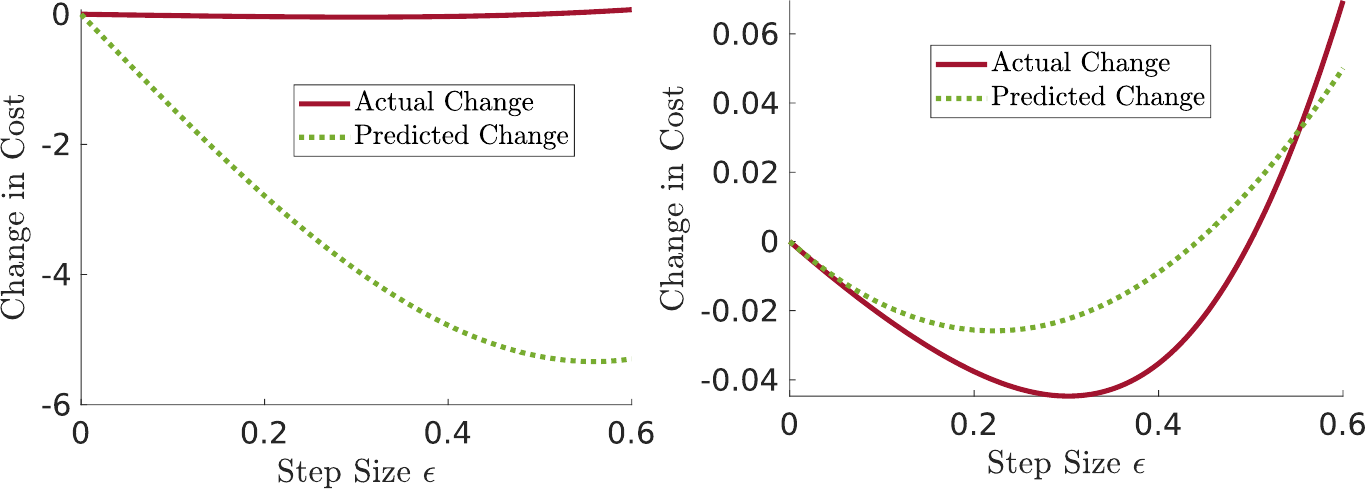}
    \caption{Expected cost change and actual cost change for the acrobot vs. step size. Left: FiLQR. Right: FiLQR with the proposed exact EC (eq.~\eqref{eq:EC}).}
    \label{fig:ls_compare}    
\end{figure}
We show the effect of the acceptance condition on the algorithm convergence using the acrobot problem. The proposed exact EC~\eqref{eq:EC} is compared against the approximated EC as used by FiLQR\cite{mastalli2020crocoddyl}. Three algorithms are evaluated, FiLQR, FiLQR-exact, and MS-iLQR. FiLQR-exact differs from FiLQR only in the expected cost change. MS-iLQR is configured to use the adaptive merit function and the exact EC~\eqref{eq:EC}, and shares everything else (backward sweep and nonlinear roll-out) in common with FiLQR. All three algorithms are configured to regard all state variables as shooting states, i.e., $M = N-1$. The initial state trajectories are obtained by linearly interpolating the initial and terminal states, while the initial controls are zero.

Figure.~\ref{fig:EC_compare} shows the comparison results in terms of cost convergence and dynamics feasibility. FiLQR fails to make progress on the given initial trajectory, whereas both the FiLQR-exact and MS-iLQR converge in 25 iterations, demonstrating the benefit of using the exact EC~\eqref{eq:EC}. To have a better understanding, the actual change in cost and the expected cost change are compared for FiLQR and FiLQR-exact. To do so, we run FiLQR-exact for several iterations, and perform a line search for both methods. The results are shown in Fig.~\ref{fig:ls_compare}. For FiLQR, the difference between the actual change and expected change is obvious, whereas for FiLQR-exact, the actual change and expected change share the same slope and concavity at step size 0, demonstrating the proposed EC is more accurate. The MS-DDP framework thus is configured to use~\eqref{eq:EC} for the rest of this work.

\subsection{Effects of Penalty Method}
\begin{figure}[t]
    \centering
    \includegraphics[width = 0.9\linewidth]{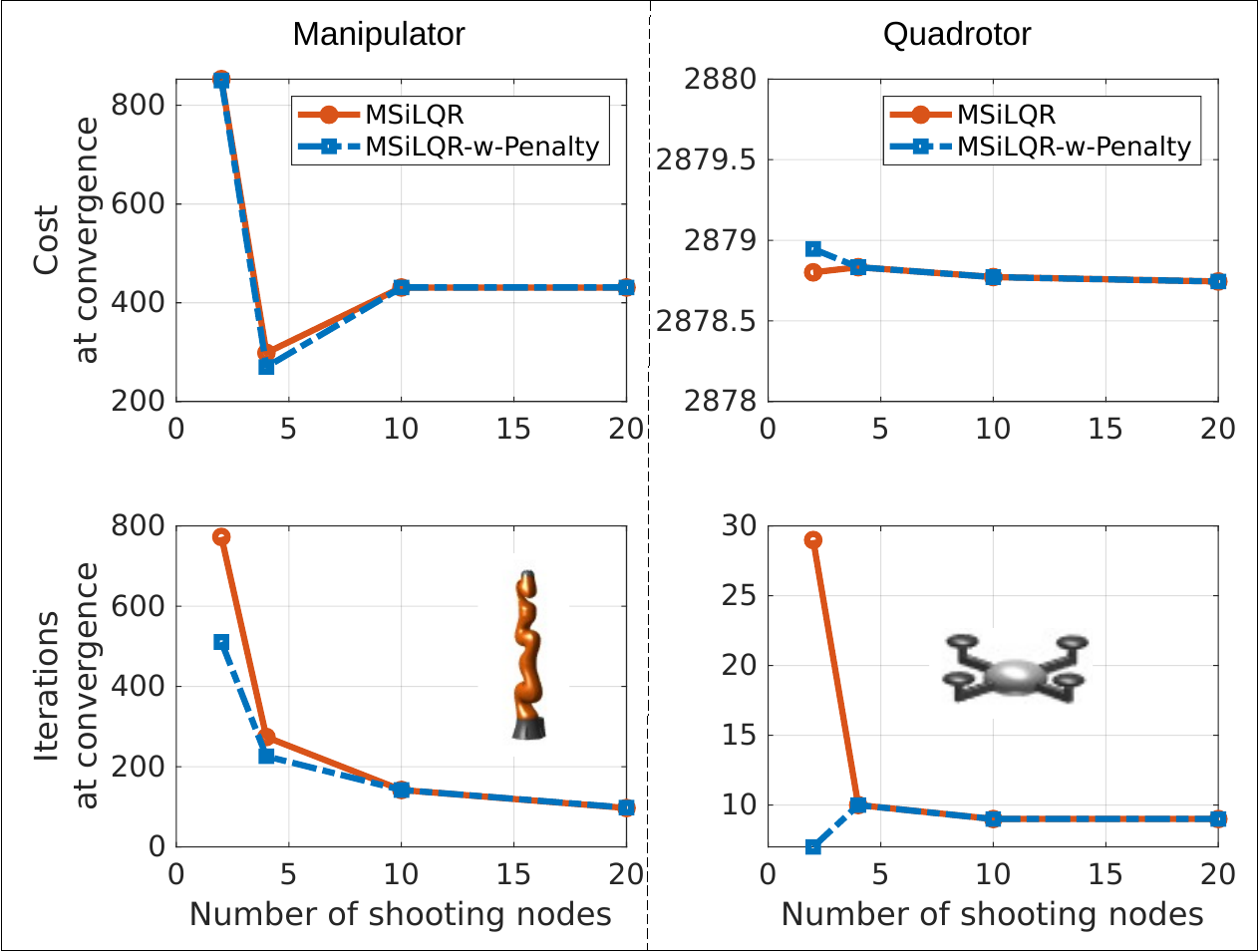}
    \caption{Effects of penalty method on algorithm convergence with a varying number of shooting nodes, benchmarked on the manipulator and quadrotor examples. Benefits are observed only for a low number of shooting nodes.}
    \label{fig:penalty_iiwa}
\end{figure}

Roughly speaking, for the multiple-shooting OCP~\eqref{eq:dt_opt}, the fewer the number of shooting segments, the more nonlinear the problem~\eqref{eq:dt_opt} is. This subsection investigates the effect of the penalty method on algorithm convergence rate for problems with different nonlinearity. To do so, we employ a varying number of shooting segments for the manipulator and the quadrotor examples. Figure~\ref{fig:penalty_iiwa} illustrates the results in terms of the cost at convergence and number of iterations to converge, acquired with and without the penalty method. MS-iLQR is configured with a hybrid roll-out and the adaptive merit function. For each example, similar costs are achieved at convergence with and without the penalty method, given the same number of shooting nodes. The penalty method, however, significantly reduces the number of iterations for the case of two shooting nodes, where higher nonlinearity arises compared to the case of more shooting nodes. Though the penalty method does not show obvious performance improvement for the quadrotor problem with four shooting nodes and above, it enables the more nonlinear manipulator problem to achieve 48 fewer iterations to converge with four shooting nodes. These observations indicate that the proposed penalty method is helpful to promote faster convergence for more nonlinear problems. Closer examinations reveal that larger step sizes are enabled with the penalty method. This result is not surprising, since both sides of the defect are considered in computing the search direction, thus facilitating faster convergence.



\begin{figure*}[!t]
    \centering
    \includegraphics[width = 0.68\linewidth]{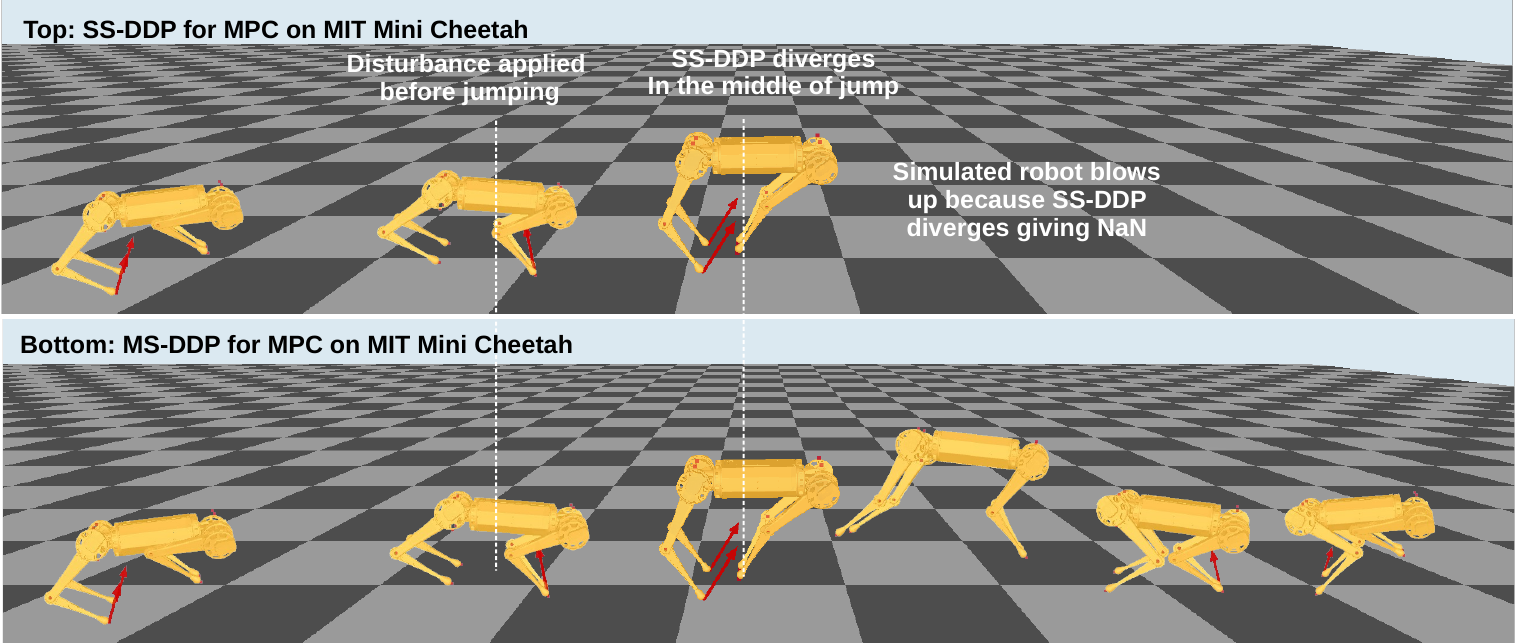}
    \caption{Snapshots of the running-to-jump gait captured in the Mini Cheetah simulator. Kick disturbance to the lateral velocity is applied before jumping. The MS-DDP (bottom) succeeds in recovering the robot from the kick, whereas the SS-DDP (top) diverges in the middle of the jump.}
    \label{fig:snapshots_run_jump}
\end{figure*}
\section{MS-DDP for MPC on A Quadruped Robot}
One motivation for this work is to develop a robust and efficient solver for real-time MPC. This section investigates the performance of the MS-DDP framework in the context of MPC for dynamic quadruped locomotion. The MPC problem is constructed based upon a hybrid kinodynamics (HKD) model \cite{li2022versatile}, which reasons about then trunk dynamics and contact-dependent leg kinematics. In previous work \cite{li2022versatile}, the HKD-MPC problem was solved using a SS-DDP variant that is tailored for hybrid systems \cite{li2020hybrid}. In this work, we adapt the previous solver to use MS-DDP, and compare its performance against the previous SS-DDP implementation.

\subsection{Simulation Results}
The comparisons are conducted on the MIT Mini Cheetah \cite{katz2019mini} in a high-fidelity simulator for multiple gaits, including mildly dynamic gaits (e.g., trotting and bounding), and highly dynamic motions (e.g., jumping). In all cases, the HKD-MPC (50 Hz) runs asynchronously from the low-level controller (500 Hz). The prediction horizon was chosen as 0.5 s with an integration time step of 10 ms. At each MPC control step, the DDP solver is terminated after two iterations. 

\begin{figure}
    \centering
    \includegraphics[width=0.9\linewidth]{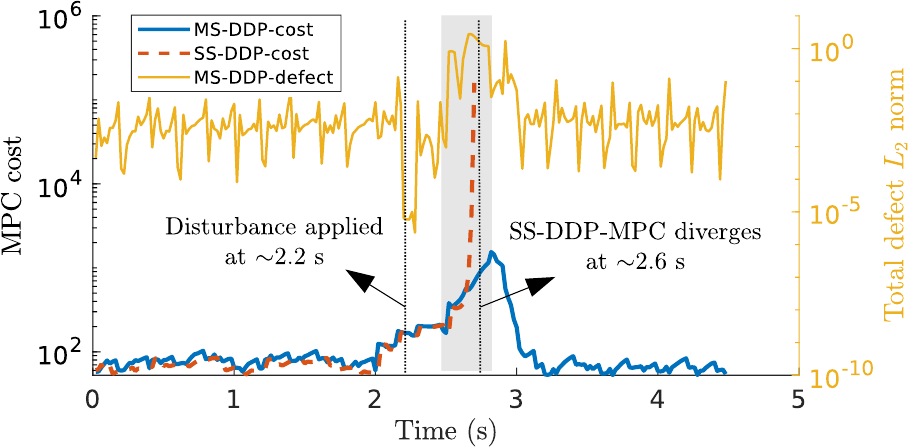}
    \caption{Accumulated cost of SS-DDP and MS-DDP, and total defect of MS-DDP at each MPC step. The defect of SS-DDP is always zero since it is inherently dynamically feasible. The grey area indicates the jumping period.}
    \label{fig:cost_compare}
\end{figure}

\begin{figure}
    \centering
    \includegraphics[width = 0.8\linewidth]{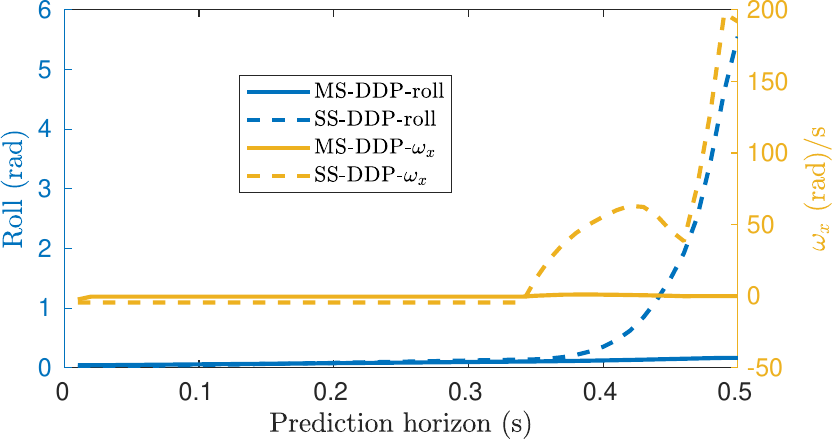}
    \caption{Nominal roll-out of SS-DDP and MS-DDP along the prediction horizon at the MPC step immediately before SS-DDP diverges. The roll angle and the angular velocity around $x-$axis are depicted.}
    \label{fig:rollout_compare}
\end{figure}

SS-DDP and MS-DDP perform equivalently well for trotting and bounding, which is reasonable since the feedback policy may be sufficient to prevent each solver from diverging for mildly-dynamic gaits. The same equivalence was not observed for the more dynamic behavior of jumping. For this motion, the robot accelerates to 2 m/s using a bounding gait, makes a jump at 2.5 s with a duration of 0.35 s, and recovers to bounding. A 0.4 m/s lateral velocity disturbance is injected before the jump. The MS-DDP enables the robot to recover stability, whereas the SS-DDP does not due to divergence in the middle of the jump. The snapshots of this result are shown in Fig.~\ref{fig:snapshots_run_jump}. To understand this difference, the accumulated costs (i.e., evaluation of the cost function~\eqref{eq:dt_cost}) are shown in Fig.~\ref{fig:cost_compare}. Both tend to increase during the jump, but SS-DDP quickly diverges. Though MS-DDP encounters a relatively-large dynamics infeasibility (i.e., total defect of 3) during the jump, this violation is temporary and is helpful to keep the accumulated cost bounded. To unveil the cause of SS-DDP divergence, we check the nominal roll-out trajectory of each algorithm at the MPC step immediately before SS-DDP diverges. Note that the roll-out trajectory is along the prediction horizon, as opposed to along the MPC step in Fig.~\ref{fig:cost_compare}. The roll rate and the roll angle are depicted in Fig.~\ref{fig:rollout_compare}. At about 0.36 s, a foot contact is established, and the nominal control policy (with feedback) of SS-DDP fails to stabilize the roll motion, thus the initial state trajectory diverges. By contrast, the MS-DDP does not suffer from this problem since its state trajectory is warm-started as well.

The results shown in Figs.~\ref{fig:snapshots_run_jump}, \ref{fig:cost_compare} and \ref{fig:rollout_compare} demonstrate the superior performance of MS-DDP over SS-DDP for highly dynamic locomotion. This conclusion is aligned with \cite{grandia2022perceptive}, which focuses on enabling the robot to traverse more complex environments using multiple shooting.

\subsection{Hardware Results}
We qualitatively validate the performance of MS-DDP on the Mini Cheetah hardware. The control setup in hardware largely matches the simulation setup with the following exceptions. First, the HKD-MPC is executed at 100 Hz on hardware. We found that 50 Hz was not sufficient to stabilize the jumping motion largely due to the increased model mismatch of the hardware. An alternative approach could be to employ a QP-based whole-body controller at a high rate. To avoid crossing singularity for the swing legs, we reduce the desired forward velocity to 1.0 $\text{m}/s$. Further, we manually push the robot to imitate the velocity disturbance. Time-series snapshots of hardware results are shown in Fig.~\ref{fig:hardware-snap}, while the complete results are in the accompanying video.
\section{Conclusions}
This work presents a unified framework for extending DDP to a multiple-shooting OCP solver. The proposed framework provides multiple configurations and several enhancements, allowing for easy comparison with and between previous algorithms. The novel derivation of the defect-aware DDP backward pass enables using second-order dynamics, and is shown to have local quadratic convergence when used with the nonlinear roll-out method. 
We show that the expected cost change model is important for algorithm convergence, and propose an exact model that further improves the performance of a state-of-the-art solver. A penalty method is introduced to provide additional robustness for problems with higher nonlinearity, and is shown to be effective in the case of a small number of shooting nodes. Future work will focus on generalizing the results to more problems with broader statistical assessment.

\section{Acknowledgements}
The Mini Cheetah is sponsored by the MIT Biomimetic Robotics Lab and NAVER LABS.

\bibliographystyle{IEEEtran}
\bibliography{ms.bib}

\end{document}